\documentclass{article} 
\usepackage{nips15submit_09,times}
\usepackage{hyperref}
\usepackage{url}

\usepackage{epsfig}
\usepackage{epstopdf}
\usepackage{graphicx}
\usepackage{color}
\usepackage{amsmath}
\usepackage{amssymb}
\usepackage[ruled,vlined,boxed]{algorithm2e}
\usepackage{multirow}
\usepackage[table]{xcolor} 
\definecolor{LightCyan}{rgb}{0.88,1,1}

\usepackage{arydshln}
\newcommand{\changeBM}[1]{#1} 
\newcommand{\changeVB}[1]{#1} 

\title{Understanding symmetries in deep networks}

\author{Vijay Badrinarayanan\\
Department of Engineering\\
Cambridge University, UK\\
\texttt{\small vb292@cam.ac.uk}\\
\And
Bamdev Mishra\thanks{This work was initiated while the author was with the Department of Electrical Engineering and Computer Science, University of Li\`ege, 4000 Li\`ege, Belgium and was visiting the Department of Engineering (Control Group), University of Cambridge, Cambridge, UK.}\\
Amazon Development Centre India\\
Bangalore, India\\
\texttt{\small bamdevm@amazon.com}\\
\And
Roberto Cipolla\\
Department of Engineering\\
Cambridge University, UK\\
\texttt{\small cipolla@cam.ac.uk}
}

\newcommand{\Diag}{{\rm Diag}}
\newcommand{\diag}{{\rm diag}}
\newcommand{\Orth}{{\rm Orth}}

\nipsfinalcopy 

\begin{document}

\maketitle

\begin{abstract}
\changeBM{Recent works} have highlighted scale invariance or symmetry present in the weight space of a typical deep network and the adverse effect it has on the Euclidean gradient based stochastic gradient descent optimization. In this work, we show that a commonly used deep network, which uses convolution, batch normalization, reLU, max-pooling, and sub-sampling pipeline, possess more complex forms of symmetry arising from scaling-based reparameterization of the network weights. \changeVB{We propose to tackle the issue of the weight space symmetry by constraining the filters to lie on the unit-norm manifold. \changeBM{Consequently, training the network boils down to using stochastic gradient descent updates on the unit-norm manifold.} Our empirical evidence based on the MNIST dataset shows that the \changeBM{proposed updates} improve the test performance beyond what is achieved with batch normalization and without sacrificing the \changeBM{computational} efficiency of the weight updates.} 

\end{abstract}

\section{Introduction}
Stochastic gradient descent (SGD) has been the workhorse for optimization of deep networks \cite{Bottou}. The most well-known form uses the Euclidean gradients with a varying learning rate to optimize the weights. In this regard, \changeBM{the} recent work \cite{PathSGD} has brought to light scale invariance properties in \changeBM{the} weight space which commonly used deep networks possess. These symmetries or \changeBM{invariance to} reparameterizations of the weights \changeBM{imply} that even though the \changeBM{loss function} remains \changeBM{unchanged}, the Euclidean gradient varies based on the chosen parameterization. \changeVB{Consequently, optimization trajectories can vary significantly for different reparameterizations \cite{PathSGD}.} 

Although these issues have been raised recently, the precursor to these methods is the early work of Amari \cite{Amari}, who proposed the use of \textit{natural gradients} to tackle weight space symmetries in neural networks. The idea is to compute the steepest descent direction for the weight update on the manifold defined by these symmetries and use this direction to update the weights \cite{PascanuNaturalGradient, DesjardinsNaturalNN, OllivierRiemmanianMetricsI, OllivierRiemmanianMetricsII}. \changeVB{Most of theses proposals are either computationally expensive to implement or they need modifications to the architecture.} \changeBM{On the other hand, optimization over a manifold with symmetries or invariances has been a topic of much research and provides guidance to other simpler metric constructions \cite{absil08a, mishra14b, boumal15a, journee10a, absil04b, edelman98a, manton02a}.}

  
\changeBM{In Section \ref{Architecture},} our analysis into a commonly used network \changeBM{shows that} there \changeBM{exists} more complex forms of symmetries which can affect optimization, and hence there is a need to define \changeBM{\emph{simpler}} weight updates which take into account these invariances. \changeBM{Accordingly, in Section \ref{Optimization}, we look at \changeBM{one particular} way of resolving the symmetries} \changeVB{by constraining the filters to lie on \changeBM{the} unit-norm manifold. This results from a geometric viewpoint on the manifold of the search space. The proposed \changeBM{updates, shown in Table \ref{ProposedUpdates}, are symmetry-invariant and are} numerically efficient to implement.}

\changeBM{The stochastic gradient descent algorithms with the proposed updates are implemented in Matlab and Manopt \cite{ManOpt}. The codes are available at \url{http://bamdevmishra.com/codes/deepnetworks}.}



\section{Architecture and symmetry analysis}
\label{Architecture}
\begin{figure*}
\centering
\includegraphics[width=0.7\textwidth]{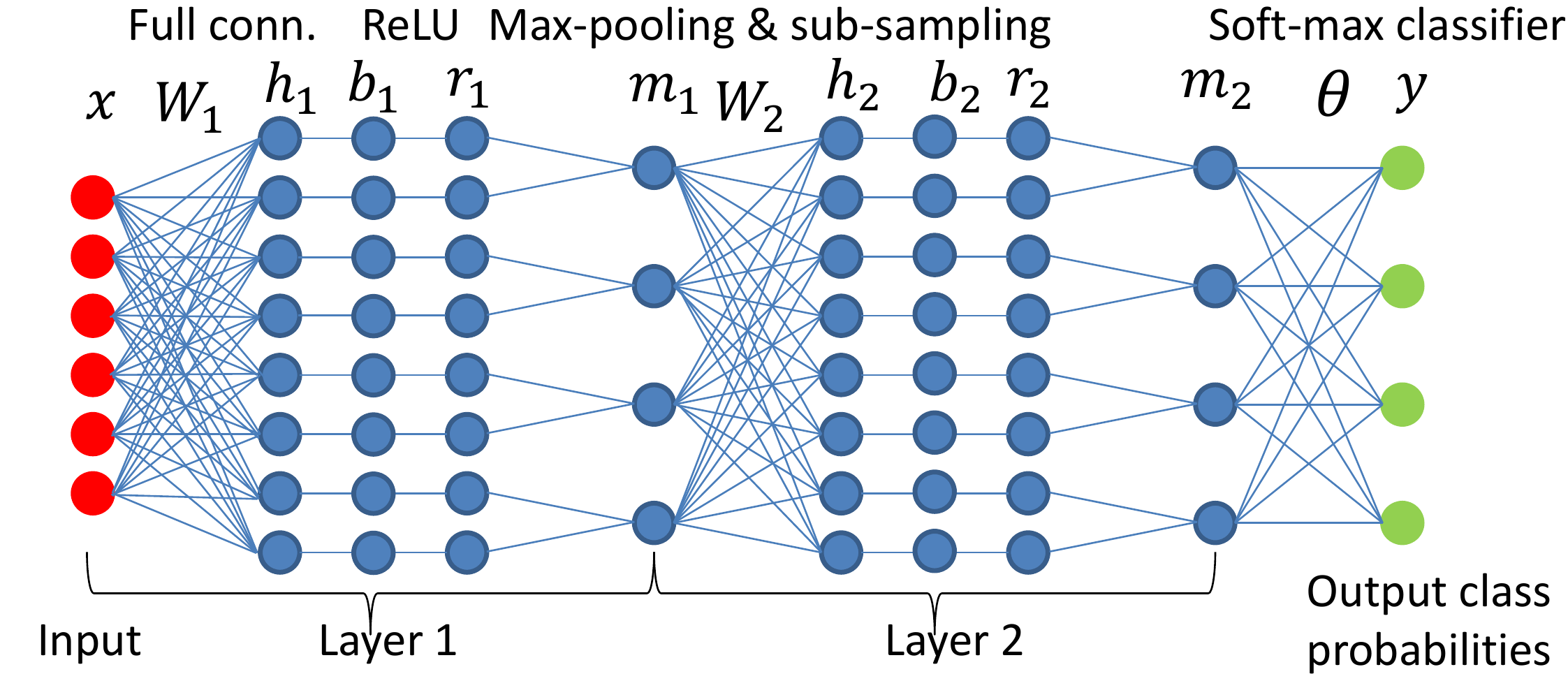}
\caption{\footnotesize{ArchBN: a two layer deep architecture for classification with batch normalization \cite{BN}.}}
\label{ArchBN}
\end{figure*}

A two layer deep architecture, ArchBN, is shown in Figure \ref{ArchBN}. Each layer in ArchBN has typical components commonly found in convolutional neural networks \cite{LeCunNature} such as multiplication with a trainable weight matrix ($W_{1},W_{2}$), a batch normalization layer ($b_{1},b_{2}$) \cite{BN}, element-wise rectification ReLU, $2\times1$ max-pooling with stride 2, and sub-sampling. The final layer is a K-way soft-max classifier $\theta$. The network is trained with a cross-entropy loss. \changeVB{The rows of the weight matrices $W_{1}$ and $W_{2}$ correspond to filters in layers $1$ and $2$, respectively. The dimension of each row corresponds to the input dimension of the layer. For the MNIST digits dataset, the input is a $784$ dimensional vector. With $64$ filters in each of the layers, the dimensionality of $W_{1}$ is $64\times784$ and of $W_{2}$ is $32\times64$. The dimension of $\theta$ is $10\times32$, where each row corresponds to a trainable class vector.} 

\changeVB{The batch normalization \cite{BN} layer normalizes each feature (element) in the $h_{1}$ and $h_{2}$ layers to have zero-mean unit variance over each mini-batch. Then a separate and trainable scale and shift is applied to the resulting features to obtain $b_{1}$ and $b_{2}$, respectively. This effectively models the distribution of the features in $h_{1}$ and $h_{2}$ as Gaussian whose mean and variance are learnt during training. Empirical results in  \cite{BN} show that this significantly improves convergence and our experiments also support this claim.} \changeBM{A key observation is that the} normalization of $h_{1}$ and $h_{2}$ allows for complex symmetries to exist in the network. \changeBM{To this end,} consider the reparameterizations
\begin{align}
\label{weightsreparamBN}
\tilde{W}_{1} =  \alpha W_{1} \quad {\rm and} \quad \tilde{W}_{2} =  \beta W_{2}, 
\end{align}
where $\mathbf{\alpha} = \Diag(\alpha_{1},\alpha_{2},\alpha_{3},\alpha_{4},\alpha_{5},\alpha_{6},\alpha_{7},\alpha_{8}) $ and $ \mathbf{\beta} = \Diag(\beta_{1},\beta_{2},\beta_{3},\beta_{4},\beta_{5},\beta_{6},\beta_{7},\beta_{8})$ and the elements of $\alpha$ and $\beta$ can be any real number. $\Diag(\cdot)$ is an operator which creates a diagonal matrix with its argument placed along the diagonal. Due to batch normalization which makes $h_1$ and $h_2$ unit-variance, $y$ is unchanged, and hence the loss is invariant to reparameterizations (\ref{weightsreparamBN}) of the weights. \changeBM{Equivalently}, there exists continuous symmetries or reparameterizations of $W_1$ and $W_2$, which leave the loss function unchanged. \changeBM{It should be stressed that our analysis differs from \cite{PathSGD}, where the authors deal with a simpler case of $\alpha = \alpha_{0}$, $\beta = \frac{1}{\alpha_{0}}$, and $\alpha_0$ is a non-zero scalar}.

\changeVB{Unfortunately, the Euclidean gradient of the weights (used in standard SGD) is not invariant to reparameterizations of the weights \cite{PathSGD}. Consequently, optimization trajectories can vary significantly based on the chosen parameterizations. This issue can be resolved either by defining a suitable non-Euclidean gradient which is invariant to reparameterizations (\ref{weightsreparamBN}) or by placing appropriate constraints on the filter weights as we show in the following section.}

\section{Resolving symmetry issues using manifold optimization}
\label{Optimization}
An efficient way to \changeBM{resolve} the symmetries that exist in ArchBN is to constrain the weight vectors (filters) in $W_{1}$ and $W_{2}$ to lie on the \emph{oblique manifold} \cite{absil08a, ManOpt}, i.e., each filter in the fully connected layers is constrained to have \changeBM{unit} norm (abbreviated UN). \changeVB{Equivalently, we impose the constraints $\diag (W_1 W_1^T) = 1$ and $\diag (W_2 W_2^T) = 1$, where $\diag(\cdot)$ is an operator which extracts the diagonal elements of the argument matrix.} To this end, consider a weight vector $w \in \mathbb{R}^{n}$ with the constraint $w^Tw = 1$. (For example, $w^T$ is a row of $W_1$.) The steepest descent direction for a loss $\ell(w)$ \changeBM{with $w$ on the unit-norm manifold} is computed $\tilde{\nabla} \ell(w) = \nabla \ell(w) - (w^{T}\nabla \ell(w)) w$, \changeBM{where $\nabla \ell(w)$ is the Euclidean gradient and $\tilde{\nabla} \ell(w)$ is the Riemannian gradient on the unit-norm manifold \cite[Chapter~3]{absil08a}.} Effectively, the normal component of the Euclidean gradient, i.e., $(w^{T}\nabla \ell(w)) w$, is subtracted to result in the tangential (to the unit-norm manifold) component. \changeBM{Following the tangential direction takes the update out of the manifold, which is then pulled back to the manifold with a \emph{retraction} operation \cite[Example~4.1.1]{absil08a}.} \changeBM{Finally}, an update of $w$ on the unit-norm manifold is of the form
\begin{equation}\label{UN}
\small{
\begin{array}{lll}
\tilde{\nabla} \ell(w) &= & \nabla \ell(w) - (w^{T}\nabla \ell(w)) w \\
\tilde{w}^{t+1}& = & w^{t} - \lambda \tilde{\nabla} \ell(w) \\ 
w^{t+1} &= & \tilde{w}^{t+1} / \| \tilde{w}^{t+1} \|, 
\end{array}
}
\end{equation}
where $w^t$ is the current weight, $w^{t+1}$ is the updated weight, $\nabla \ell(w)$ is the Euclidean gradient, and $\lambda$ is the learning rate. \changeBM{It should be noted that when $W_1$ and $W_2$ are constrained, the $\theta$ variable is reparameterization free.}

\changeVB{The proposed weight update (\ref{UN}) can be used in a stochastic gradient descent (SGD) setting which we use in our experiments described in the following section. \changeBM{It should be emphasized that the proposed update is numerically efficient to implement. The formulas are shown in Table \ref{ProposedUpdates}. The convergence analysis of SGD on manifolds follows the developments in \cite{Bottou, bonnabel13a}.}}

\begin{table*}[t]%
\center \scriptsize 
\begin{tabular}{|c c|}
\hline
$\begin{array}[t]{lll}
\tilde{{\nabla}}_{W_1} L (W_1^t, W_2^t, \theta^t) = \Pi_{W_1^t}({{\nabla}}_{W_1} L (W_1^t, W_2^t, \theta^t))\\
W_1^{t+1} = {\Orth}({W_1^t} - \lambda \tilde{{\nabla}}_{W_1} L (W_1^t, W_2^t, \theta^t))
\end{array} $ 
&
 $\begin{array}[t]{lll}
\tilde{{\nabla}}_{W_2} L (W_1^t, W_2^t, \theta^t) = \Pi_{W_2^t}({{\nabla}}_{W_2} L (W_1^t, W_2^t, \theta^t))\\
W_2^{t+1} = {\Orth}({W_2^t} - \lambda \tilde{{\nabla}}_{W_1} L (W_1^t, W_2^t, \theta^t))
\end{array} $  \\

& \\
\multicolumn{2}{|c|}{
$\begin{array}[t]{lll}
\theta^{t+1} = \theta^t - \lambda  {\nabla}_{\theta} L (W_1^t, W_2^t, \theta^t) 
\end{array} $ } \\

\hline
\end{tabular}%
\caption{\footnotesize{The proposed UN symmetry-invariant updates for a loss function $L(W_1, W_2, \theta)$ in ArchBN. Here $(W_1^t, W_2^t ,\theta_t)$ is the current weight, $(W_1^{t+1} , W_2^{t+1}, \theta^{t+1})$ is the updated weight, $\lambda$ is the learning rate, and ${\nabla}_{W_1} L (W_1^t, W_2^t,\theta^t)$,  ${\nabla}_{W_2} L (W_1^t, W_2^t, \theta^t)$, and ${\nabla}_{\theta} L (W_1^t, W_2^t, \theta^t)$ are the partial derivatives of the loss $L$ with respect to $W_1$, $W_2$, and $\theta$, respectively at $(W_1^t, W_2^t, \theta ^t)$. The operator $\Orth(\cdot)$ normalizes the rows of the input argument. $\Pi_W(\cdot)$ is the linear projection operation that projects an arbitrary matrix onto the tangent space of the oblique manifold at an element $W$. It is defined as $\Pi_W(Z) = Z - \Diag(\diag((ZW^T))W$ \cite{ManOpt}.}}
\label{ProposedUpdates}
\end{table*}

\section{Experiments and results}
\label{Experiments}
We train both two and four layer deep ArchBN to perform digit classification on the MNIST dataset ($60$K training and $10$K testing images). We use 64 features per layer. The digit images are rasterized into a $784$ dimensional vector as input to the network(s). No input pre-processing \changeBM{is} performed. The weights in each layer are drawn from a standard Gaussian and each filter is unit-normalized. The class vectors are also drawn from a standard Gaussian and unit-normalized.  


We use SGD-based optimization and use choose the base learning rate from the set $10^{-p}$ \changeBM{for $p \in \{2, 3, 4, 5\}$} for each training run. For finding the base learning rate, we create a validation set of $500$ images from the training set. We then train the network with a fixed learning rate using a randomly chosen set of $1000$ images for $50$ epochs. At the start of each epoch, the training set is randomly permuted and mini-batches are sampled in a sequence ensuring each training sample is used only once within an epoch. We record the validation error measured as the error per training sample for each candidate base learning rate. We then choose the candidate rate which corresponds to the lowest validation error and use this for training the network on the full training set. We repeat this whole process for $10$ training runs for each network to measure the mean and variance of the test error. \changeBM{We ignore the runs where the validation error diverged.} \changeVB{For each full dataset training run, we use the bold-driver protocol \cite{HintonBoldDriver} to anneal the learning rate.} We choose $50000$ randomly chosen samples as the training set and the remaining $10000$ samples for validation. We train for a minimum of $25$ epochs and a maximum of $60$ epochs. Training is terminated if either the training error is less than $10^{-5}$ or the validation error increases with respect to the one measured before $5$ epochs or successive validation error measurements differ less than $10^{-5}$.

\begin{table*}[t]%
\centering 
\scriptsize{
\begin{tabular}{|c|c|c|c|}
\hline
Depth & B-SGD &  UN \\
\hline
2 &  0.0206 $\pm$ 0.0024 &  \textbf{0.0199} $\pm$ 0.0046 \\
\hline
4 & 0.0204 $\pm$ 0.0027 & \textbf{0.0179} $\pm$ 0.0025 \\
\hline
\end{tabular}}%
\caption{\footnotesize{The test error after the last training epoch measured over $10$ runs on the MNIST dataset. \changeBM{The proposed UN update shows a competitive performance while resolving the symmetries.}} }
\label{TestErr}
\end{table*}

\changeVB{
From Table \ref{TestErr} we see that for both two and four layer deep networks, the mean test error is lower for UN as compared to balanced SGD (B-SGD) which is simply the Euclidean update, but where the starting values of filters and class vectors are unit-normalized. The lowest mean and variance in the test error is obtained when UN weight update is used for training a four layer deep network. The difference between the B-SGD and UN updates is more significant for the four layer deep network, thereby highlighting the performance improvement over what is achieved by \changeBM{standard} batch normalization in deeper networks. The use UN can also be seen as \changeBM{a} way to regularize the weights of the network during training without introducing any hyper-parameters, \changeBM{e.g., a weight decay term}.} It should also be noted that the performance difference between the two and four layer networks is not very large. This raises the question for future research as to whether some deep networks necessarily have to be that deep or it can be made shallower (and efficient) by better optimization \cite{Caruana}. 

%


\begin{figure*}[t]
\centering
\includegraphics[width=0.6\textwidth]{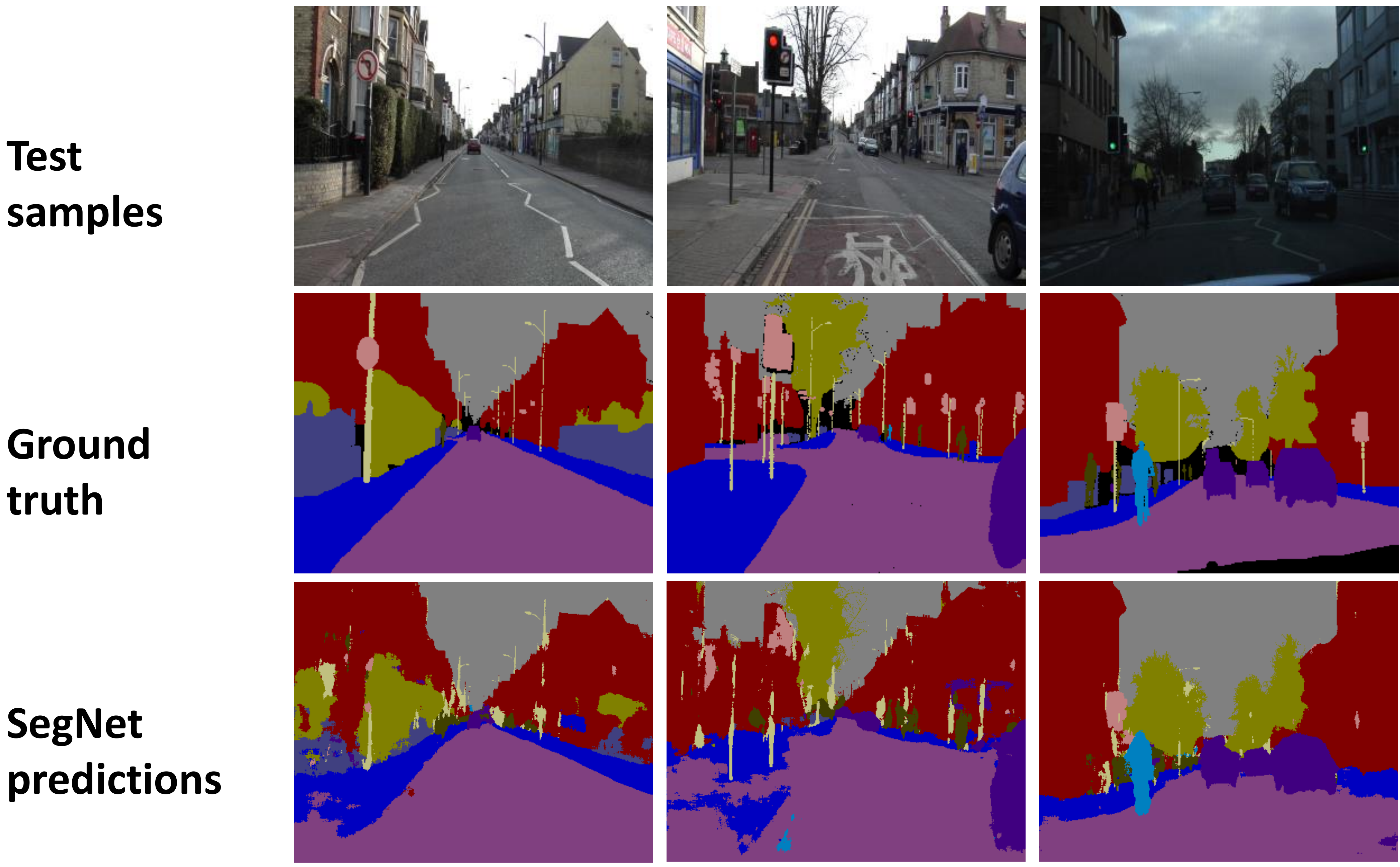}
\caption{\footnotesize{SGD with the proposed UN weight updates, shown in Table \ref{ProposedUpdates}, for training SegNet \cite{SegNetarXiv}. The quality of the predictions as compared to the ground truth indicates a successful training of SegNet.}}
\label{CamVidQualy}
\end{figure*}

\section{Application to image segmentation}
\label{DeepConvNets}
We \changeBM{apply SGD with the proposed UN weight updates in Table \ref{ProposedUpdates} for} training SegNet, a deep convolutional network proposed for road scene image segmentation into multiple classes \cite{SegNetarXiv}. This network, although convolutional, possesses the same symmetries as those analyzed for ArchBN in (\ref{weightsreparamBN}). The network \changeBM{is} trained for 100 epochs on the CamVid \cite{GabeDataset} training set of 367 images. The predictions on some sample test images from  CamVid are shown in Figure \ref{CamVidQualy}. These qualitative results indicate the usefulness of \changeBM{symmetry-invariant} weight updates for larger networks that arise in practice. 


%
%


\section{Conclusion}
\label{Conclusion}
We have highlighted the symmetries that exist in the weight space of deep neural network architectures which are currently popular. \changeVB{These symmetries can be absorbed into gradient descent by applying a unit-norm constraint on the filter weights. This takes into account the manifold structure on which the weights of the network reside. The empirical results show the test performance can be improved using our proposed weight update technique on a modern architecture}. \changeBM{As a future research direction,} we would \changeVB{like to explore other efficient symmetry-invariant weight update techniques and exploit them for deep convolutional neural network \changeBM{used in} practical applications}.

\subsubsection*{Acknowledgments}
Bamdev Mishra was supported as an FNRS research fellow (Belgian Fund for Scientific Research). The scientific responsibility rests with its authors.

\small
\bibliographystyle{unsrt}  
\bibliography{optarXiv_BMC}


%

\end{document}